\documentclass[runningheads]{llncs}

\usepackage{eccv}

\usepackage{eccvabbrv}

\usepackage{graphicx}
\usepackage{booktabs}

\usepackage{amsmath}
\usepackage{amssymb}
\usepackage{booktabs}
\usepackage{adjustbox}	
\usepackage{color, colortbl}
\usepackage{multirow}
\usepackage{makecell}
\usepackage{pifont}
\usepackage{tikz,pgfplots}

\usepackage[accsupp]{axessibility}  
\newcommand\para[1]{\vspace{3mm}\noindent\textbf{#1}}

\usepackage[pagebackref,breaklinks,colorlinks,citecolor=eccvblue]{hyperref}

\usepackage{orcidlink}

\begin{document}
	
	\title{ProDepth: Boosting Self-Supervised Multi-Frame Monocular Depth with Probabilistic Fusion}

	\titlerunning{ProDepth}
	
	\author{Sungmin Woo\inst{*} \and
		Wonjoon Lee\inst{*} \and
		Woo Jin Kim \and
		Dogyoon Lee \and
		Sangyoun Lee}
	
	\authorrunning{S.~Woo et al.}
	
	\institute{Yonsei University \\
		\url{https://sungmin-woo.github.io/prodepth/}}
	\maketitle
	
	\def\thefootnote{*}\footnotetext{co-first authors}
	
	\begin{abstract}
		Self-supervised multi-frame monocular depth estimation relies on the geometric consistency between successive frames under the assumption of a static scene. However, the presence of moving objects in dynamic scenes introduces inevitable inconsistencies, causing misaligned multi-frame feature matching and misleading self-supervision during training. In this paper, we propose a novel framework called ProDepth, which effectively addresses the mismatch problem caused by dynamic objects using a probabilistic approach. We initially deduce the uncertainty associated with static scene assumption by adopting an auxiliary decoder. This decoder analyzes inconsistencies embedded in the cost volume, inferring the probability of areas being dynamic. We then directly rectify the erroneous cost volume for dynamic areas through a Probabilistic Cost Volume Modulation (PCVM) module. Specifically, we derive probability distributions of depth candidates from both single-frame and multi-frame cues, modulating the cost volume by adaptively fusing those distributions based on the inferred uncertainty. Additionally, we present a self-supervision loss reweighting strategy that not only masks out incorrect supervision with high uncertainty but also mitigates the risks in remaining possible dynamic areas in accordance with the probability. Our proposed method excels over state-of-the-art approaches in all metrics on both Cityscapes and KITTI datasets, and demonstrates superior generalization ability on the Waymo Open dataset. 
	\end{abstract}
	
	\begin{figure*}[t]
		\centering
		\includegraphics[width=\linewidth]{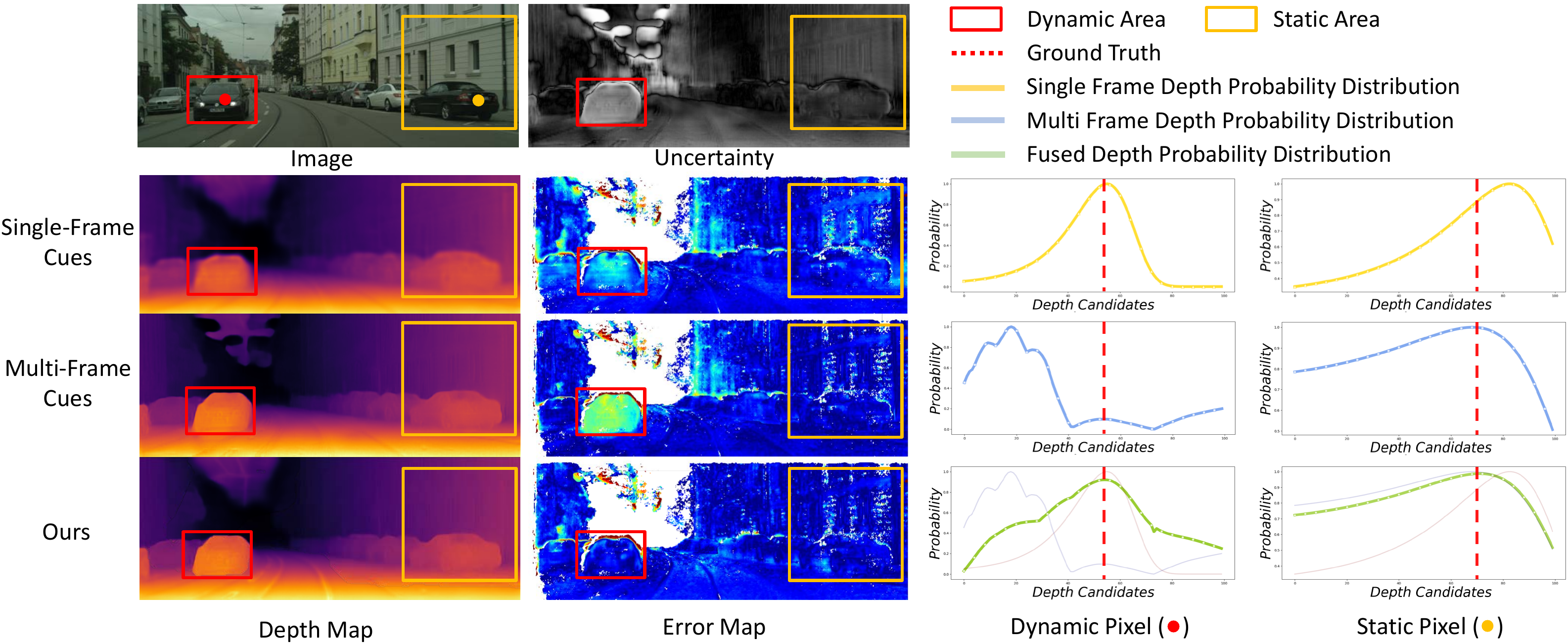}
		
		\caption{Our ProDepth performs uncertainty-aware adaptive fusion of the probability distributions from both single-frame and multi-frame cues. The fused distribution follows the distribution of single-frame cues for a dynamic pixel, while adhering to the distribution of multi-frame cues for a static pixel. Error maps in the second column depict large depth errors in green and small in blue.}
		
		\label{fig:1}
	\end{figure*}

	\section{Introduction}
	\label{sec:intro}
	
	Accurate depth information is essential across various domains, including autonomous driving, robotics, and augmented reality. The deployment of precise 3D sensors (\eg, structured light or LiDAR) is often hindered by their high costs, leading to the development of depth estimation solely from RGB images. Notably, self-supervised monocular depth estimation from single or multiple frames is gaining traction, removing the need for ground-truth data from costly sensors.
	
	Early self-supervised depth estimation methods~\cite{bian2019unsupervised, chen2019self, monodepth, vijayanarasimhan2017sfm, zhou2017unsupervised} take a single target image to infer depth by analyzing visual patterns including texture, shading, and edges. The adjacent images are only incorporated for self-supervision at the training-level, which is achieved by minimizing a photometric reprojection error~\cite{zhou2017unsupervised} between frames as a novel-view synthesis problem. However, due to the constraints of limited information, their performance falls short of achieving satisfactory results. Recently, multi-frame based approaches~\cite{manydepth, depthformer, dynamicdepth} have emerged to leverage temporally adjacent frames as valuable geometric cues for depth estimation. These methods perform multi-frame feature matching within the cost volume under the assumption of a static scene, assessing the probabilities of various depth candidates for each pixel based on geometric consistency between frames. Despite their overall high performance, these approaches exhibit significant errors in dynamic areas. The inconsistent geometric locations of moving objects lead to misaligned feature matching, resulting in an incorrect depth probability distribution.
	
	To address the mismatch problem in the cost volume, several works~\cite{manydepth, depthformer, dynamicdepth} leverage single-frame depth to compensate for errors in dynamic areas in multi-frame depth. The underlying insight~\cite{li2023learning, monorec} is that multi-frame based estimation tends to yield more accurate predictions in static areas, whereas single-frame based estimation without the cost volume avoids misaligned feature matching, thereby better handling moving objects (Fig.~\ref{fig:1}). The representative approach~\cite{manydepth, depthformer} is to supervise dynamic areas of multi-frame depth with single-frame depth by using an additional training loss term, aiming to enforce correct depth estimation despite an incorrect cost volume. However, this loss-level solution cannot entirely prevent errors in the cost volume from affecting the final prediction, as the fundamental issue of an incorrect multi-frame matching cost distribution persists. Recently, DynamicDepth~\cite{dynamicdepth} indirectly addresses the mismatch issue in the cost volume by adjusting the locations of dynamic objects in input images to be static using single-frame depth. However, this process requires pre-computed segmentation masks to identify objects, and the masks involve needless static objects since segmentation does not account for their movements. While these approaches have made progress in handling dynamic areas, their limitations highlight the need for further exploration. Our key observation is that accurately identifying dynamic objects remains a significant challenge, and the direct refinement of incorrect matching costs in the cost volume has yet to be thoroughly explored.

	In this paper, we introduce ProDepth, a novel framework that makes three major contributions to address the inconsistency issue caused by dynamic objects. First, rather than relying on additional semantic information, we discern an uncertainty map (\textit{i.e.}, the probability that each pixel is not static) using an auxiliary depth decoder. This decoder deliberately predicts corrupted depth based on erroneous cost volume, enabling to infer object-level uncertainty from the extent of corruption. Second, we present a Probabilistic Cost Volume Modulation (PCVM) module, which directly rectifies the erroneous matching costs of the cost volume through uncertainty-aware adaptive fusion of single- and multi-frame cues. As illustrated in Fig.~\ref{fig:1}, the depth probability distribution adaptively follows either the distribution of single-frame cues or that of multi-frame cues based on the inferred uncertainty. Finally, to further mitigate the issues associated with incorrect self-supervision of reprojection loss in dynamic areas during training, we devise a loss reweighting strategy. This strategy entails adjusting the computed reprojection loss according to uncertainty, thereby reducing incorrect supervision in possible dynamic areas. In summary, we present the following noteworthy contributions:
	
	\begin{itemize}
		\item We devise an auxiliary depth decoder, which facilitates the identification of moving objects as a probabilistic representation, \textit{i.e.} uncertainty, without using of a pretrained off-the-shelf segmentation network.
		\item We propose PCVM, a novel approach addressing the mismatch problem in the cost volume by directly rectifying the corrupted matching cost distribution through the probabilistic fusion of single-frame and multi-frame cues.
		\item We introduce a self-supervision loss reweighting strategy to counteract incorrect supervision in potential dynamic areas, distinct from conventional binary masking methods.
		\item Our approach achieves state-of-the-art results on Cityscapes and KITTI datasets, and also demonstrates superior generalization ability on the Waymo Open dataset.
	\end{itemize}

	\section{Related Work}
	\label{sec:rel}
	\subsection{Self-Supervised Monocular Depth Estimation}
	Conventional single-frame based methods use a single image for estimation, with temporally adjacent images employed solely for self-supervision during training. A self-supervised framework~\cite{zhou2017unsupervised} is proposed to compute photometric consistency between monocular frames, facilitating joint training of a single-frame depth estimation network and a multi-frame camera pose estimation network. Subsequent advancements are achieved in camera geometry modeling~\cite{featdepth, gordon2019depth}, network architectures~\cite{packnet, litemono}, reprojection loss~\cite{zhanst2018, featdepth} and the handling of depth errors in moving objects~\cite{ranjan2018adversarial, monodepth2, gordon2019depth, vijayanarasimhan2017sfm, geonet2018, chen2019self, bian2019unsupervised, sgdepth, li2021unsupervised}. Recent approaches~\cite{manydepth, monorec, dynamicdepth, depthformer, dualrefine, wang2020self, ruhkamp2021attention, patil2020dont, Casser_2019_CVPR_Workshops} have shifted towards integrating temporal information not only in the training loss function but also in depth prediction. The current state-of-the-art methods~\cite{depthformer, manydepth, monorec, dynamicdepth} adopt the cost volume generally used in stereo matching tasks to capture geometric compatibility between images. As a pioneering work, ManyDepth~\cite{manydepth} introduces an adaptive cost volume to overcome the scale ambiguity problem in self-supervised monocular depth estimation. To enhance multi-frame feature matching in the cost volume, DepthFormer~\cite{depthformer} incorporates attention mechanisms, replacing conventional similarity metrics with a learnable matching function. Building on these works, we also utilize a multi-frame cost volume but effectively address the misaligned feature matching problem caused by dynamic objects through probabilistic cost volume modulation.
	
	\subsection{Dynamic Objects in Static Scene Constraint}
	As homography warping is employed in cost volume construction and photometric reprojection loss based on the assumption of a static scene, the presence of moving objects inevitably causes incorrect matching costs and misleading supervision. To tackle the inherent challenges of multi-view inconsistency for dynamic objects in self-supervised depth learning, two key steps are essential: (1) dynamic objects should be identified from the rigid background, and (2) errors in cost volume and reprojection loss must be rectified.

	\noindent\textbf{Discerning dynamic objects.}\hspace{0.3cm}
	To identify dynamic areas, a typical approach is to use a pretrained semantic~\cite{sgdepth, guizilini2020semantically, dynamicdepth} network or an instance segmentation network~\cite{instadm, Casser_2019_CVPR_Workshops, casser2019depth, monorec}. While leveraging an useful off-the-shelf network is effective in discerning moving objects, it comes with several drawbacks, including an added computational burden, the potential inclusion of static objects in segmentation masks, and confinement to predefined classes. In contrast, our proposed ProDepth identifies potential moving objects solely based on the provided images, eliminating the requirement for additional information.

	\noindent\textbf{Rectifying errors caused by dynamic objects.}\hspace{0.3cm}
	As discussed in Sec.~\ref{sec:intro}, recent multi-frame based methods address the mismatch problem in the cost volume through indirect approaches, such as supervising predicted depth with single-frame depth for potential dynamic areas at the loss-level~\cite{manydepth, depthformer} or removing the motion of dynamic objects at the input-level~\cite{dynamicdepth}. In contrast, our ProDepth aims to directly rectify the erroneous matching costs with the proposed Probabilistic Cost Volume Module (PCVM), performing motion-aware adaptive fusion of single-frame and multi-frame cues in a probabilistic manner. Additionally, to tackle the incorrect supervision in dynamic areas, existing methods~\cite{dynamicdepth,depthformer,manydepth,litemono, monodepth2} use a binary mask to exclude the computed loss in those regions. However, binary masking for the estimated moving objects may not adequately consider possible dynamic areas involving ambiguous probability. Instead, we propose a loss reweighting strategy, which partially reduces incorrect supervision based on the inferred probability.

	\section{Method}
	\label{sec:method}
	
	\subsection{Self-Supervised Monocular Depth Learning}
	\label{sec:ssl}
	Given the target image $I_t$ and temporally adjacent source images $\{I_s\mid s\in\{t-1,t+1\}\}$, we can warp $I_s$ to the view point of $I_{t}$ with the estimated depth of the target image $D_t$ and the relative camera pose $T_{t\rightarrow s}$:
	\begin{equation}
		\label{eq:1}
		I_{s\rightarrow t}(D_t) = I_s\big\langle\text{proj}(D_t,T_{t\rightarrow s},K)\big\rangle,
	\end{equation}
	where $K$ is the known camera intrinsics, $\text{proj}(\cdot)$ indicates the projection of 3D points from $D_t$ into the camera of $I_s$, and $\big\langle\cdot\big\rangle$ is the pixel sampling operator. For self-supervised learning of depth and camera ego-motion, the photometric reprojection loss is generally used for optimization that consists of structure similarity (SSIM)~\cite{ssim} and $\textit{L}_1$ loss terms:
	\begin{equation}
		\mathcal{L}_{p}(D_t) = \alpha~\frac{1-\text{SSIM}(I_t,I_{s\rightarrow t}(D_t))}{2}+(1-\alpha)~\| I_t - I_{s\rightarrow t}(D_t) \|_1,
	\end{equation}
	where $\alpha$ is commonly set to 0.85. Importantly, this reprojection loss provides misleading supervision for dynamic areas because the image warping process is based on the static scene assumption.
	
	\begin{figure*}[t]
		\centering
		\includegraphics[width=\linewidth]{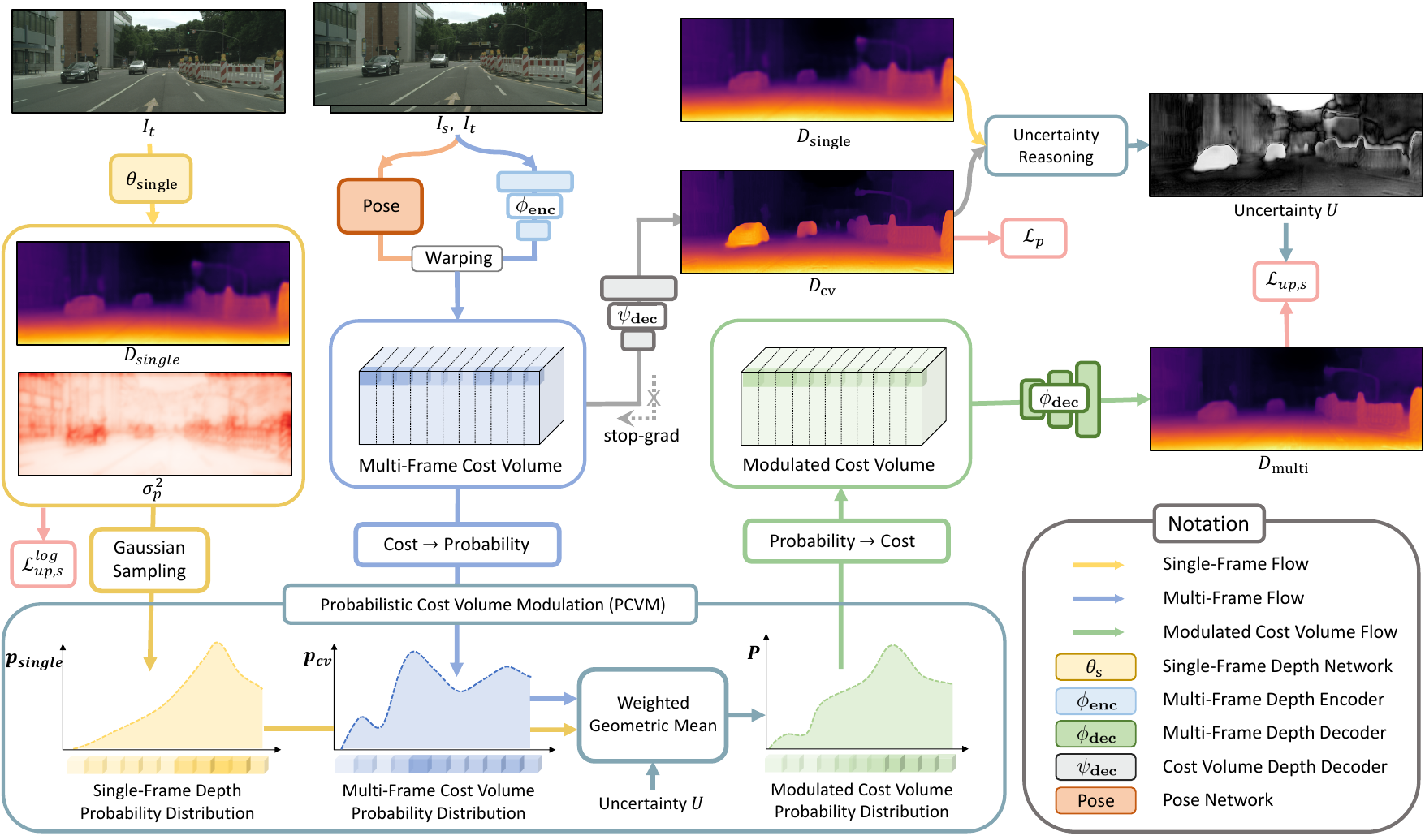}
		
		\caption{\textbf{Overview of the proposed ProDepth.} We construct the multi-frame cost volume with $I_{s}$ and $I_t$, and estimate single-frame depth as a Gaussian distribution using the target image $I_t$. In an auxiliary branch, uncertainty is inferred by comparing $D_{\text{single}}$ and $D_\text{cv}$, where the latter is estimated from cost volume features. To rectify erroneous cost volume, a PCVM module adaptively fuses probabilities derived from single- and multi-frame cues. Furthermore, we incorporate a loss reweighting strategy in $\mathcal{L}_{up,s}$ and $\mathcal{L}^{log}_{up,s}$ to mitigate errors caused by moving objects at the training-level. Note that the probability distribution of a dynamic pixel is illustrated as an example.}
		\label{fig:2}
	\end{figure*}
	
	\subsection{Overview}
	The proposed architecture contains three major components that address the inconsistency issue caused by moving objects. Initially, we identify uncertainty by analyzing depth maps estimated from auxiliary depth decoders (Sec.~\ref{sec:2}). Subsequently, a PCVM module rectifies erroneous matching costs in the cost volume for dynamic areas by uncertainty-aware adaptive fusion of probability distributions of depth candidates from single- and multi-frame cues (Sec.~\ref{sec:3}). At the training-level, we mitigate misleading self-supervision devising a loss reweighting strategy (Sec.~\ref{sec:4}). The overall framework is summarized in Fig.~\ref{fig:2}.
	
	\subsection{Auxiliary Depth Estimations and Uncertainty Reasoning}
	\label{sec:2}
	To reason uncertainty associated with the static scene assumption and compensate for errors in dynamic areas in multi-frame cost volume, our framework incorporates two auxiliary depth estimations: single-frame depth and cost volume depth.

	\noindent\textbf{Probabilistic single-frame depth estimation.}\hspace{0.3cm}
	We employ a lightweight network, denoted as $\theta_\text{single}$, to estimate single-frame depth $D_\text{single} \in \mathbb{R}^{H\times W}$ using a target image $I_t \in \mathbb{R}^{H\times W}$. To estimate the depth as a probability distribution, we adopt the predictive approach~\cite{kendall2017uncertainties, klodt2018supervising, poggi2020uncertainty}, configuring the network to output the mean $\mu$ and variance $\sigma^2$ of the distribution in the final layer. Specifically, we model the predictive distribution as a heteroscedastic Gaussian, minimizing the negative log-likelihood criterion. For supervised learning with ground-truth depth $D^*$, the negative log-likelihood is given by:
	\begin{equation}
		-\text{log}~p(D^*|\mu, \sigma) = \frac{(D^*-\mu)^2}{\sigma^2}+ \text{log}~\sigma^2.
	\end{equation}
	In our self-supervised learning scenario, where ground-truth $D^*$ is unavailable, we predict the variance map $\sigma_p^2  \in \mathbb{R}^{H\times W}$ for pixel-wise photometric matching between the target image and warped image as shown in \cite{klodt2018supervising, poggi2020uncertainty}:
	\begin{equation}
		\label{eq:4}
		\mathcal{L}_p^\text{log}(D_{\text{single}}) = \frac{(\mathcal{L}_{p}(D_{\text{single}}))^2}{\sigma^2_{p}}+ \text{log}~\sigma^2_p.
	\end{equation}
	Through log-likelihood maximization of $\mathcal{L}_p^\text{log}$, we estimate single-frame depth as a probability distribution with mean $D_{\text{single}}$ and variance $\sigma^2_p$.
	
	\begin{figure*}[t]
		\centering
		\includegraphics[width=\linewidth]{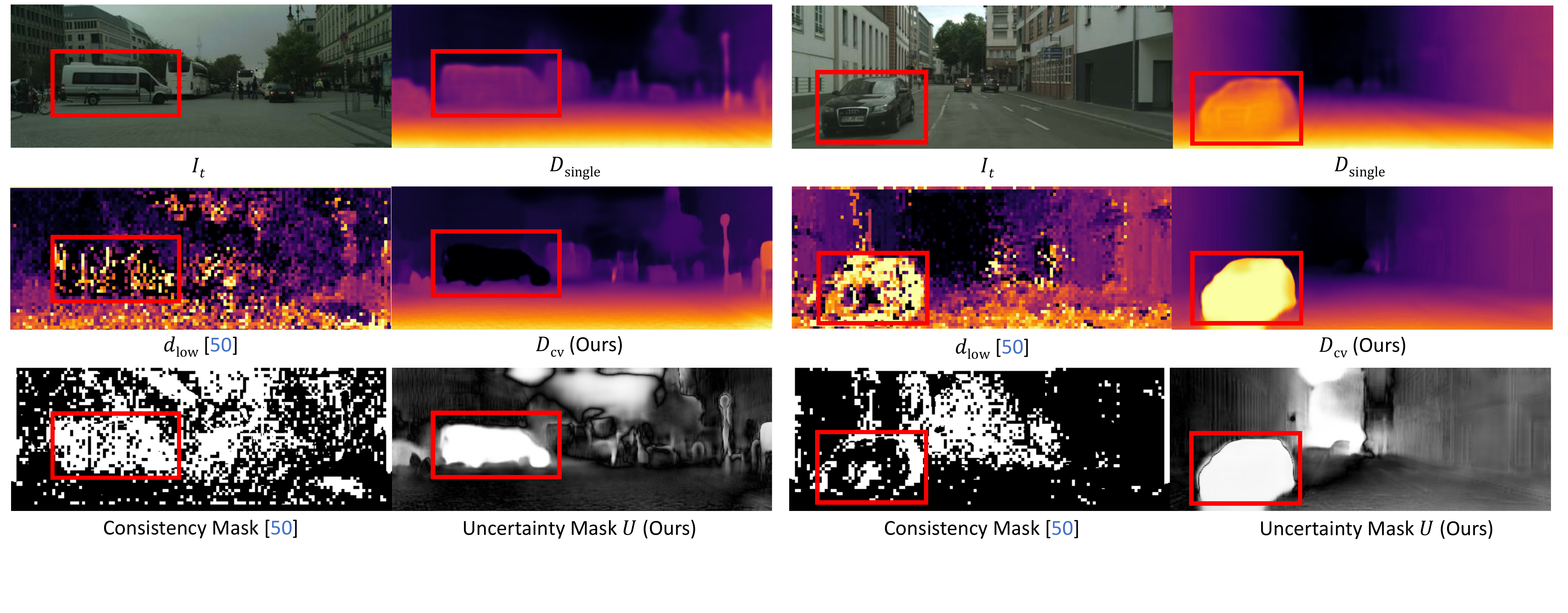}
		\vspace{-1cm}
		\caption{\textbf{The identification of dynamic objects.} In contrast to the binary consistency mask generated in ManyDepth~\cite{manydepth}, our uncertainty reasons the probability of moving objects with structural awareness.}
		\label{fig:3}
	\end{figure*}

	\noindent\textbf{Cost volume depth estimation and uncertainty reasoning.}\hspace{0.3cm}
	In multi-frame depth encoder $\phi_{\text{enc}}$, we first encode $I_t$ and $I_s$ into $C$-dimensional downsampled features $F_t$ and $F_s$ with size $H/4 \times W/4 \times C$. We then construct the cost volume $\mathcal{C}$ to measure the multi-frame matching costs for hypothesized depth candidates $d=\{d_i\mid i\in\{1,2,...,k\}\}$. Depth candidates are perpendicular to the optic axis of $I_t$ and uniformly sampled in log space by spatial-increasing discretization~\cite{fu2018deep}:
	\begin{equation}
		d_i = e^{\text{log}(d_1)+\frac{i}{k-1}\text{log}(d_k/d_1)},
	\end{equation}
	where depth candidates range from $d_1$ to $d_k$, representing the minimum and maximum depth values, respectively. For each depth candidate $d_i$, the source feature $F_s$ is warped to the view point of $I_t$, producing $F_{s\rightarrow t}(d_i)$ similar to $I_{s\rightarrow t}(D_t)$ in Eq.~\ref{eq:1}. We compute the per-pixel matching costs for all $d_i$ by the absolute $\textit{L}_1$ difference between $F_t$ and $F_{s\rightarrow t}$, and aggregate feature channels by average pooling to obtain the cost volume $\mathcal{C} \in \mathbb{R}^{H/4 \times W/4 \times k}$. The cost is expected to be lower for the depth candidate that is closer to the ground-truth depth.
	
	As the cost volume construction involves the static scene assumption in the warping process of $F_{s\rightarrow t}(d_i)$, the inconsistent geometric locations of moving objects result in misaligned feature matching with incorrectly computed matching costs. The corrupted cost distribution in dynamic areas then leads to erroneous depth estimation, degrading overall performance. However, this corruption can be leveraged to identify moving objects by comparing with accurately predicted single-frame depth. ManyDepth~\cite{manydepth} generates a binary mask called \textit{consistency mask} where single-frame depth $D_{\text{single}}$ and $\textit{argmin}$ of the matching costs $d_{\text{low}}$ (\textit{i.e.}, $d_i$ with lowest cost) differ significantly, considering it as unreliable region involving high uncertainty. The problem with this approach is that the mask relying on lowest cost cannot clearly mask out moving objects because $d_{\text{low}}$ is computed for each pixel independently based on feature distance without an understanding of spatial correlation between pixels, \textit{i.e.} structural awareness.

	To overcome these limitations, we devise an auxiliary decoder $\psi_{\text{dec}}$ that estimates the depth using corrupted matching costs of the cost volume. Our main observation is that depth estimation goes beyond capturing pixel-level geometric information; it integrates structural awareness, ensuring that pixels within the same object demonstrate consistent depth values. Decoding pixel-level inconsistencies embedded in the cost volume into depth $D_{\text{cv}}$ results in consistent errors within moving objects, as shown in Fig.~\ref{fig:3}. Based on the generated clear corruption, we discern the uncertainty $U \in[0,1]^{H\times W}$ by computing the absolute difference between $D_{\text{single}}$ and $D_{\text{cv}}$ and normalizing it into $[0,1]$ with the mapping function $\mathcal{M}(a, b)= 1 - e^{-\beta|a-b|}$:
	
	\begin{equation}
		U = \mathcal{M}(D_{\text{single}}, D_{\text{cv}}) = 1 - e^{-\beta|D_{\text{single}} - D_{\text{cv}}|},
	\end{equation}
	where $\beta$ is empirically set to 0.6. Unlike the obscure binary mask~\cite{manydepth} generated from roughly computed per-pixel lowest cost, our uncertainty precisely indicates the probability of object-level corruption, which can be utilized as an useful cue for identifying dynamic objects.

	\subsection{Probabilistic Cost Volume Modulation}
	\label{sec:3}
	Contrary to existing works, we address the errors embedded in the cost volume by directly modulating the matching cost distribution. We first transform the single- and multi-frame cues into a representation of probability distribution along depth candidates and derive the modulated cost distribution for each pixel by adaptively fusing those distributions based on the uncertainty.

	\noindent\textbf{Single-frame depth as probability distribution.}\hspace{0.3cm}As shown in Eq.~\ref{eq:4}, we estimate the mean $D_\text{single}$ and variance $\sigma_p^2$ of the single-frame depth as Gaussian distribution. By using the probability density function of Gaussian distribution $\mathcal{N}(D_\text{single}, \sigma_p^2)$, we can compute the probability of each depth candidate $d_i$ given pixel $x$:
	\begin{equation}
		p_{\text{single}}(d_i|x) = \frac{1}{\sqrt{2\pi \sigma_p^2(x)}}\exp(-\frac{(d_i-D_\text{single}(x))^2}{2\sigma_p^2(x)}),
	\end{equation}
	where $D_\text{single}(x)$ and $\sigma_p^2(x)$ indicate the estimated mean and variance values for pixel $x$, respectively.

	\noindent\textbf{Multi-frame matching costs as probability distribution.}\hspace{0.3cm}For the initially constructed cost volume $\mathcal{C} \in \mathbb{R}^{H/4 \times W/4 \times k}$, we denote the matching cost of depth candidate $d_i$ for the given pixel $x$ as $\mathcal{C}(x,i)$. The per-pixel costs are converted into probabilities $p_{\text{cv}}(d|x)$ using the softmax function:
	\begin{equation}
		p_{\text{cv}}(d_i|x) = \frac{\exp(-\mathcal{C}(x,i))}{\sum_{j=1}^{k}\exp(-\mathcal{C}(x,j))},
	\end{equation}
	where negative costs are used for softmax because the depth candidate with lower cost holds a higher probability.

	\noindent\textbf{Cost volume modulation.}\hspace{0.3cm}
	\label{cv}
	The modulation of cost volume involves the fusion of the probability distributions of depth candidates derived from single-frame depth and multi-frame cost volume considering the uncertainty. To preserve the relative importance of each distributions after fusion, we adopt weighted geometric mean (Eq.~\ref{eq:wgm}) which allows for the retention of depth candidates with highest probability due to its multiplicative nature. In contrast, the commonly used weighted arithmetic mean (weighted sum) with additive nature may not guarantee the preservation of depth candidates at the maximum due to the linear combination of distributions. We present an ablation study of the fusion strategy in the supplementary material.
	
	Based on the computation of weighted geometric mean, probabilities $p_j \in \{p_{\text{single}}, p_{\text{cv}}\}$ are multiplied with consideration for each weight $w_j \in \{U, 1-U\}$ to derive the fused probability distribution $P(d|x)$ (Eq.~\ref{eq:fusion}):
	\begin{align}
		\label{eq:wgm}
		P(d|x) & = (\prod_{j}{p_j(d|x)^{w_j}})^{1/{\sum_{j}{w_j}}} \\ \label{eq:fusion}
		& = p_{\text{single}}(d|x)^{U(x)} \cdot p_{\text{cv}}(d|x)^{1-U(x)}.
	\end{align}
	For non-static pixels with high uncertainty, the distribution of single-frame depth $p_\text{single}$ exerts a greater influence, whereas for static pixels with low uncertainty, the distribution of the cost volume $p_\text{cv}$ carries more weight. The fused probability distribution $P$ is then re-scaled to the scale of original cost volume $\mathcal{C}$ by min-max normalization to obtain modulated cost volume $\mathcal{C}_{m}$: 
	\begin{equation}
		\mathcal{C}_{m}(x,i) = \frac{\text{max}(P(d|x)) - P(d_i|x)}{\text{max}(P(d|x)) - \text{min}(P(d|x))}(\text{max}(\mathcal{C}(x)) - \text{min}(\mathcal{C}(x))) + \text{min}(\mathcal{C}(x)),
	\end{equation}
	where $\frac{\text{max}(P(d|x)) - P(d_i|x)}{\text{max}(P(d|x)) - \text{min}(P(d|x))}$ inverts the fused probability distribution of depth candidates while normalizing to $[0, 1]$, since lower matching cost indicates higher probability in the original cost volume. The final multi-frame depth $D_{\text{multi}}$ is subsequently estimated from $\mathcal{C}_{m}$ using the decoder $\phi_{\text{dec}}$, wherein errors caused by dynamic objects are rectified through the cost volume modulation.

	\subsection{Learning without Dynamic Objects}
	\label{sec:4}
	
	\noindent\textbf{Uncertainty-aware loss reweighting strategy.}\hspace{0.3cm}
	As addressed in Sec.~\ref{sec:ssl}, an optimization of photometric reprojection loss $\mathcal{L}_{p}$ provides incorrect supervision for non-static pixels. To exclude the misleading loss for dynamic areas, we devise a loss reweighting strategy that adjusts the computed reprojection loss based on the uncertainty. The uncertainty-aware photometric reprojection loss $\mathcal{L}_{up}$ is formulated as:
	\begin{equation}
		\mathcal{L}_{up} = M\odot(1-U)\odot\mathcal{L}_{p},\hspace{0.3cm}M = [U < \gamma],
	\end{equation}
	where $\odot$ is element-wise product and $[\cdot]$ denotes the Iverson bracket. The computed $\mathcal{L}_{p}$ is reweighted based on the per-pixel probability of uncertainty $U\in[0,1]^{H\times W}$, and additional binary mask $M$ is applied to rigorously exclude pixels involving high uncertainty. In comparison to conventional binary masking methods employed in existing works, our loss reweighting strategy is more effective in preventing erroneous depth overfitting for moving objects. This effectiveness stems from its ability to partially reduce incorrect supervision for areas with ambiguous uncertainty, which may not be adequately addressed by a binary mask. Combining both binary masking and probabilistic reweighting allows us to mitigate the risk in defining learning objectives for potential dynamic areas while unequivocally excluding incorrect supervision associated with high uncertainty.

	\noindent\textbf{Objective functions.}\hspace{0.3cm}
	Continuing with the established procedure~\cite{manydepth, dynamicdepth, depthformer}, we incorporate the edge-aware smoothness loss $\mathcal{L}_{s}$~\cite{monodepth} to regularize the smoothness of the predicted depth map, and consistency loss $\mathcal{L}_{c}~$\cite{manydepth, dynamicdepth, depthformer} to ensure multi-frame depth to be similar to single-frame depth in dynamic areas. 
	
	We denote $\mathcal{L}_{up,s} = \mathcal{L}_{up} + \lambda_{s}\mathcal{L}_{s}$ as a combination of uncertainty-aware photometric reprojection loss and smoothness loss, and our final loss $\mathcal{L}$ is
	\begin{equation}
		\mathcal{L}= \underset{x}{\sum}~[ \mathcal{L}_{up,s}(D_{\text{multi}}) + \lambda_1\mathcal{L}^{log}_{up,s}(D_{\text{single}}) + \lambda_2\mathcal{L}_{p}(D_\text{cv}) + \lambda_3\mathcal{L}_c],
	\end{equation}
	where $x$ indicates pixel index. For multi- and single-frame depth estimation, our uncertainty-aware reprojection loss $\mathcal{L}_{up}$ is employed to prevent erroneous overfitting for moving objects. In contrast, $\mathcal{L}_{p}$ is used in cost volume depth estimation to encourage corruption, enabling the identification of dynamic regions with moving objects through significant depth difference between $D_\text{cv}$ and $D_\text{single}$. By allowing incorrect self-supervision in dynamic areas, cost volume decoder $\psi_{dec}$ learns to produce erroneous depth in moving objects. Note that for $\mathcal{L}_{p}(D_\text{cv})$, backpropagation is enabled exclusively for the parameters of the cost volume decoder, while the gradients are halted from flowing through the cost volume. 
	
	\section{Experiments}
	\label{sec:exp}
	We evaluate the performance of our approach on two challenging datasets, Cityscapes~\cite{Cityscapes} and KITTI~\cite{kitti}, recognized benchmarks for depth estimation. Since Cityscapes dataset contains more moving objects compared to KITTI, our experiments are mainly focused on Cityscapes to verify the performance improvement in dynamic scenes. We conduct quantitative and qualitative comparisons with state-of-the-art methods, and an extensive ablation study to substantiate the contributions of the proposed components. Given the importance of evaluating performance in dynamic regions for our work, additional experimental results can be found in the supplementary material.
	
	\subsection{Experimental Setup}
	\noindent\textbf{Dataset.} In our study of the Cityscapes dataset, we use a set of pre-processed 58,335 training images provided by \cite{dynamicdepth}, along with 1,525 images for testing. For the KITTI dataset, we adhere to the Eigen split~\cite{eigen-split} following established practices~\cite{manydepth, depthformer, dynamicdepth, dualrefine}. This split encompasses 39,810 training images, 4,424 validation images, and 697 test images. In both datasets, we exclusively use unlabeled video frames, without incorporating additional segmentation masks or optical flow information. The ground-truth depth information is employed solely for evaluation, and we constrain the predicted depth values to be below 80 meters.

	\para{Metrics.}
	We evaluate the depth performance using widely adopted metrics~\cite{eigen-split}, including four error metrics (Abs Rel, Sq Rel, RMSE, and RMSE log) and three accuracy metrics ($\delta < 1.25$, $\delta < 1.25^2$, and $\delta < 1.25^3$).
	
	\begin{table*}[t!]
		\centering
		\caption{Depth evaluation on the Cityscapes and KITTI datasets.  
			\textit{Semantics} indicates the use of additional semantic information.}
		\label{tab:main}
		\resizebox{\textwidth}{!}{%
			\begin{tabular}{c|l|ccc|ccccccc}
				\Xhline{4\arrayrulewidth}
				\multirow{2}{*}{}&\multirow{2}{*}{Method} & \multirow{2}{*}{Test frames} & \multirow{2}{*}{Semantics} & \multirow{2}{*}{$W\times H$} & \multicolumn{4}{c}{\cellcolor{red!7}Error metric ($\downarrow$)}    & \multicolumn{3}{c}{ \cellcolor{green!7}Accuracy metric ($\uparrow$)}  \\
				& & & & & 
				\cellcolor{red!7} Abs Rel & \cellcolor{red!7} Sq Rel&  \cellcolor{red!7} RMSE  & \cellcolor{red!7} RMSE log & \cellcolor{green!7} $\delta < 1.25 $ & \cellcolor{green!7} $\delta < 1.25^2$  & \cellcolor{green!7} $\delta < 1.25^3$     \\
				\hline
				\multirow{15}{*}{ \rotatebox{90}{KITTI}} 
				&Struct2depth (M)~\cite{casser2019depth} & 1 & \ding{51} & $416\times 128$ & 0.141 & 1.026 & 5.291 & 0.215 & 0.816 & 0.945 & 0.979 \\
				
				&Videos in the wild~\cite{gordon2019depth} & 1 & \ding{51} & $416\times 128$ & 0.128 & 0.959 & 5.230 & 0.212 & 0.845 & 0.947 & 0.976 \\

				&Johnston \etal~\cite{johnston2020self} & 1 & & $640\times 192$ & 0.111 & 0.941 & 4.817 & 0.189 & 0.885 & 0.961 & 0.981\\
				
				&Packnet-SFM~\cite{packnet} & 1 & & $640\times 192$ & 0.111 & 0.785 & 4.601 & 0.189 & 0.878 & 0.960 & 0.982 \\
				
				&Monodepth2~\cite{monodepth2} & 1 & & $640\times 192$ & 0.110 & 0.831 & 4.642 & 0.187 & 0.883 & 0.962 & 0.982\\
				
				&HR-Depth~\cite{hrdepth} & 1 & & $640\times 192$ & 0.109 & 0.792 & 4.632 & 0.185 & 0.884 & 0.962 & 0.983\\
				
				&Guizilini \etal~\cite{guizilini2020semantically} & 1 & \ding{51} & $640\times 192$ & 0.102 & 0.698 & 4.381 & 0.178 & 0.896 & 0.964 & 0.984 \\
				
				&Lite-Mono~\cite{litemono} & 1 & & $640\times 192$ & 0.101 & 0.729 & 4.454 & 0.178 & 0.897 & 0.965 & 0.983 \\								\arrayrulecolor{gray}\cline{2-12}
				&Patil \etal~\cite{patil2020dont} & N & & $640\times 192$ & 0.111 & 0.821 & 4.650 & 0.187 & 0.883 & 0.961 & 0.982\\
				&Wang \etal~\cite{wang2020self} & 2 (-1, 0) & & $640\times 192$ & 0.106 & 0.799 & 4.662 & 0.187 & 0.889 & 0.961 & 0.982 \\				
				&ManyDepth~\cite{manydepth} & 2 (-1, 0) & & $640\times 192$ & 0.098 & 0.770 & 4.459 & 0.176 & 0.900 & 0.965 & \underbar{0.983} \\
				&DynamicDepth~\cite{dynamicdepth} & 2 (-1, 0) & \ding{51} & $640\times 192$ &0.096 & 0.720 & 4.458 &0.175 &0.897 &0.964& \textbf{0.984} \\
				&DepthFormer~\cite{depthformer} & 2 (-1, 0) & & $640\times 192$ & \underbar{0.090} & 0.661 & \underbar{4.149} & 0.175 & 0.905 & \underbar{0.967} &\textbf{0.984} \\
				& DualRefine~\cite{dualrefine} & 2 (-1, 0) & & $640\times 192$ & \underbar{0.090} & \underbar{0.658} & 4.237 & \underbar{0.171} & \underbar{0.912} & \underbar{0.967} & \textbf{0.984} \\
				&\cellcolor{blue!7}
				\textbf{ProDepth} & 2 (-1, 0) & & $640\times 192$ & \textbf{0.086} & \textbf{0.629 }& \textbf{4.139}& \textbf{0.166}& \textbf{0.918}& \textbf{0.969}& \textbf{0.984} \\
				\Xhline{4\arrayrulewidth}
				
				\multirow{11}{*}{ \rotatebox{90}{Cityscapes}}
				&Pilzer \etal \cite{pilzer2018unsupervised} & 1 &  & $512\times256 $&0.240 & 4.264 & 8.049 & 0.334 & 0.710 &  0.871 & 0.937 \\
				&Struct2Depth 2 \cite{Casser_2019_CVPR_Workshops}  &  1  & &  $416\times128$ &0.145  & 1.737  & 7.280  &  0.205 & 0.813 & 0.942 & 0.976 \\
				
				&Monodepth2 \cite{monodepth2} & 1  && $416\times128$ &0.129  &   1.569  &   6.876  &   0.187  &   0.849  &   0.957  &   0.983 \\
				
				&Videos in the Wild \cite{gordon2019depth} & 1 & & $416\times128$ &{0.127} & {1.330} & {6.960} & {0.195} & {0.830} & {0.947} & {0.981} \\
				&Li \etal \cite{li2020unsupervised} & 1 & &$416\times128$ &0.119 &  1.290 & {6.980} &  {0.190} & {0.846} &  0.952 &  0.982 \\
				& Lee \etal~\cite{lee2021attentive} & 1 & &$832\times256$ &0.116 &1.213 &6.695 &0.186 &0.852 &0.951 &0.982\\
				& InstaDM~\cite{instadm}  & 1     & \ding{51}& $832\times256$ &0.111 &1.158 &6.437 &0.182 &0.868 &0.961 &0.983 \\
				\arrayrulecolor{gray}\cline{2-12}
				&Struct2Depth 2 \cite{Casser_2019_CVPR_Workshops}  &  3 (-1,  0,  +1) & \ding{51} & $416\times128$ &0.151 & 2.492 & 7.024 & 0.202 & 0.826 & 0.937 & 0.972 \\
				&ManyDepth~\cite{manydepth} & 2 (-1, 0)  & & $416\times128$ &0.114  &   1.193  &   6.223  &   0.170  &   0.875  &   0.967  &   0.989 \\
				&DynamicDepth~\cite{dynamicdepth} & 2 (-1, 0)   & \ding{51} &$416\times128$ &\underbar{0.103} &\underbar{1.000} &\underbar{5.867} &\underbar{0.157} &\underbar{0.895} &\underbar{0.974} &\underbar{0.991}  \\
				&\cellcolor{blue!7}
				\textbf{ProDepth} & 2 (-1, 0) & &$ 416\times128$ & \textbf{0.095}  & \textbf{0.876}& \textbf{5.531}& \textbf{0.146} & \textbf{0.908}& \textbf{0.978}& \textbf{0.993} \\
				\Xhline{4\arrayrulewidth}
			\end{tabular}%
		}
	\end{table*}
	
	\begin{figure*}[t]
		\centering  
		\includegraphics[width=0.9\linewidth]{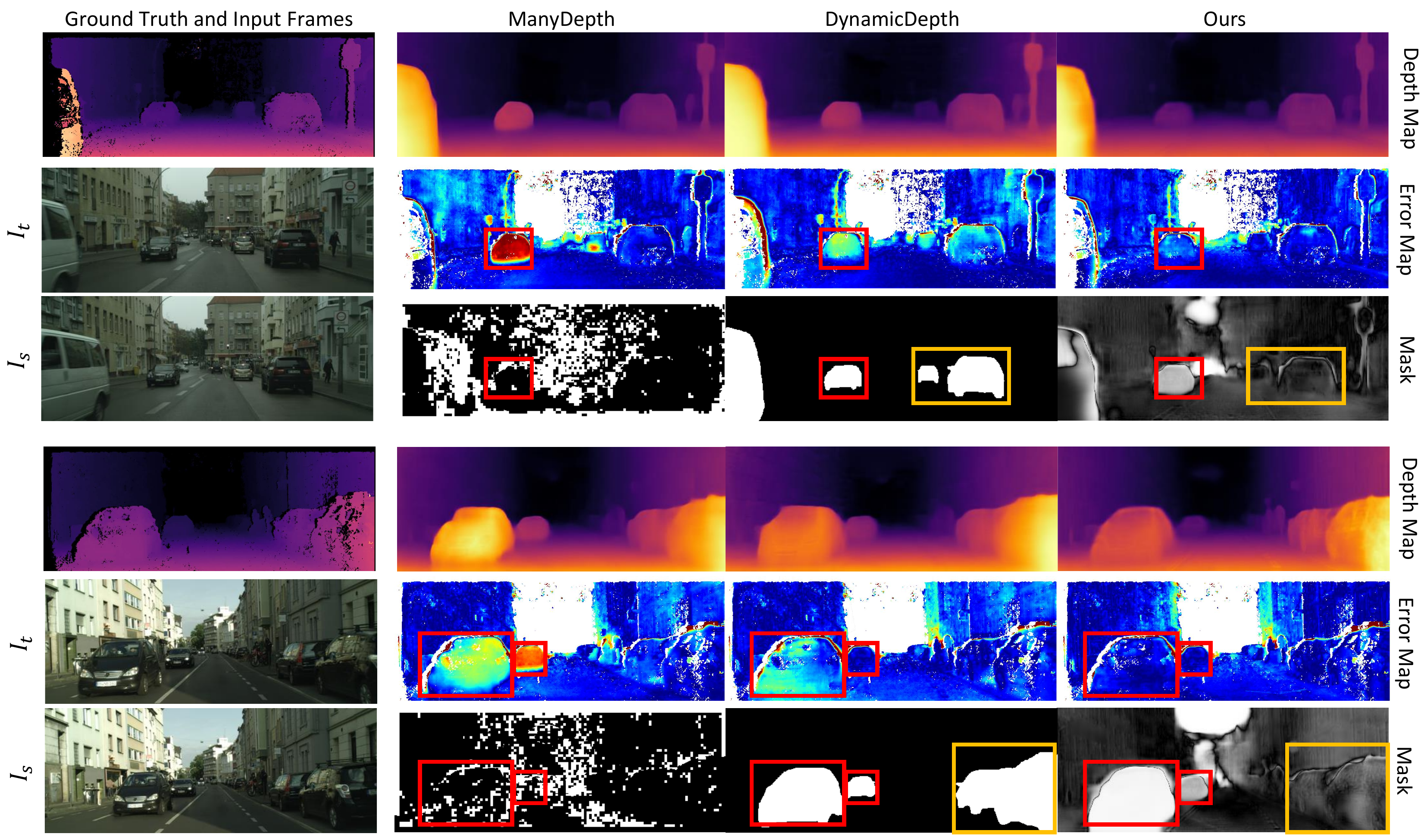}
		\caption{\textbf{Qualitative results on Cityscapes.} Red and yellow boxes indicate moving and static objects. Error maps depict large depth errors in red and small in blue.}
		\label{fig:4}
	\end{figure*}

	\subsection{Results on Cityscapes}
	Table~\ref{tab:main} presents a performance comparison between our approach and state-of-the-art methods on the Cityscapes~\cite{Cityscapes} and KITTI~\cite{kitti} datasets. Notably, for the Cityscapes dataset, which includes a significant number of moving objects, our proposed ProDepth achieves a remarkable improvement over existing methods across all metrics. It is worth highlighting that ProDepth, relying solely on the given input images, outperforms approaches~\cite{dynamicdepth, Casser_2019_CVPR_Workshops, instadm} that utilize additional semantic information. This underscores the effectiveness of our uncertainty reasoning in discerning dynamic objects. Additionally, we present qualitative results on the Cityscapes test set in Fig.~\ref{fig:4}. While related works~\cite{manydepth, dynamicdepth} exhibit relatively high estimation errors in dynamic areas, our ProDepth demonstrates superior performance. 
	
	\subsection{Results on KITTI}
	We further evaluate our proposed ProDepth on the KITTI dataset using the Eigen split. According to the statistics analyzed in \cite{dynamicdepth}, the pixels indicating movable objects in dynamic classes constitute 0.34\% of all pixels. As static objects are considered together in the statistics, KITTI involves a fewer dynamic areas compared to Cityscapes. Nevertheless, our model still outperforms recent works including both single-frame and multi-frame based approaches. This demonstrates that our probabilistic fusion of single-frame and multi-frame cues also benefits the prediction in static scenes.
	
	\subsection{Ablation Study}
	In Table~\ref{tab:ablation}, we conduct extensive ablation study to evaluate three major contributions: (1) uncertainty reasoning with an auxiliary depth decoder, (2) a probabilistic cost volume modulation (PCVM) module, and (3) an uncertainty-aware loss reweighting strategy.
	
	\noindent\textbf{Uncertainty reasoning.}\hspace{0.3cm}
	The identification of dynamic areas can be represented in a binary or weighted (probabilistic) manner. ManyDepth~\cite{manydepth} adopts a binary consistency mask estimated at the coarse feature-level, while
	DynamicDepth~\cite{dynamicdepth} employs a pretrained semantic segmentation network to identify the movable objects, as shown in Fig.~\ref{fig:4}. To assess the effectiveness of each mask, we substitute our uncertainty reasoning with their masks in our model during both training and inference (row \#1$\sim$4). Additionally, we convert our weighted uncertainty into a binary representation by setting a threshold (row \#5,6). Our model using binary uncertainty shows similar performance to the model using segmentation masks, demonstrating that our auxiliary decoder discerns the moving objects effectively. It is noteworthy that our uncertainty does not indicate a high probability in static objects, unlike segmentation masks of DynamicDepth~\cite{dynamicdepth}.
	
	\begin{figure*}[t]
		\centering  
		\includegraphics[width=0.9\linewidth]{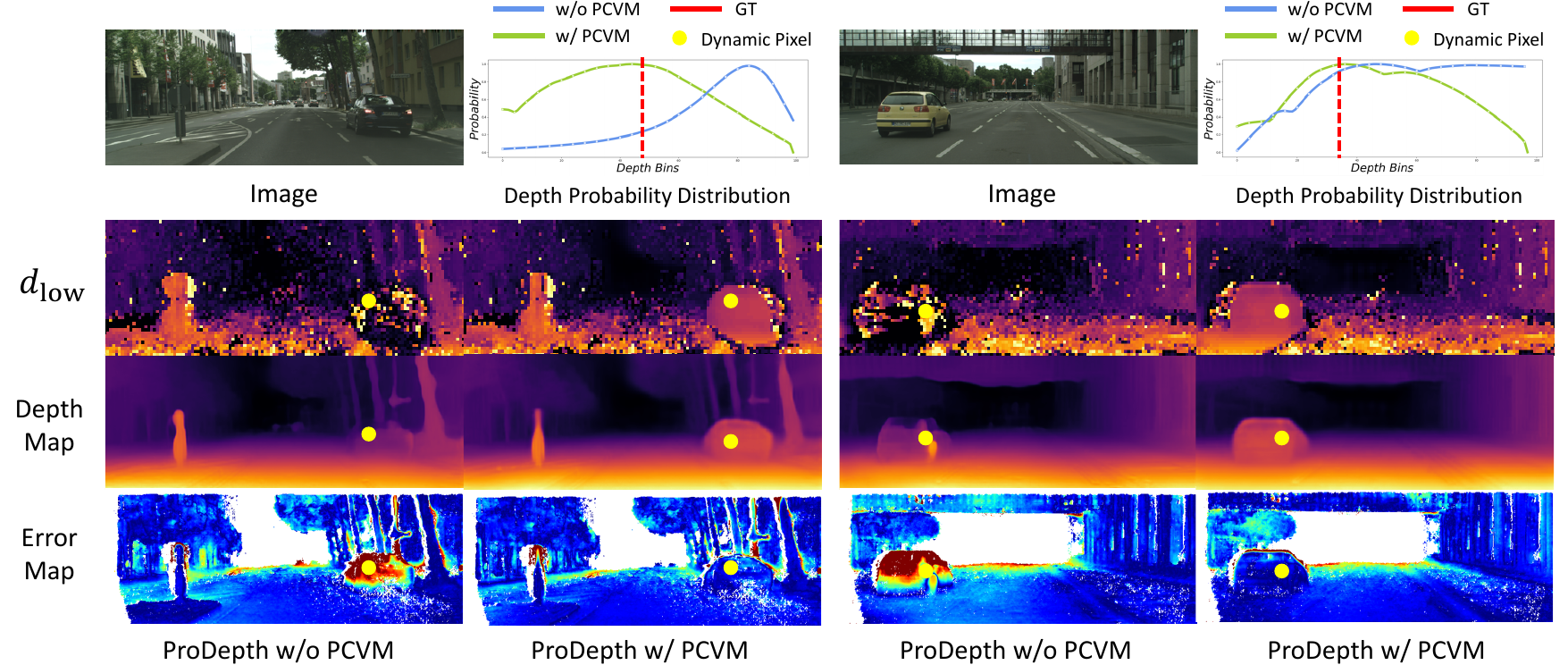}
		\caption{\textbf{ProDepth with and without the PCVM module.} Depth probability distributions of a dynamic yellow pixel are presented. Our PCVM modulates the incorrect distribution in cost volume, rectifying the errors in dynamic areas. }
		\label{fig:5}
	\end{figure*}
	
	\begin{table}[t!]
		\centering
		\caption{Ablation study on the Cityscapes dataset.}
		\label{tab:ablation}
		\resizebox{\textwidth}{!}{%
			\begin{tabular}{c|>{\centering}p{3.4cm}>{\centering}p{2cm}|>{\centering}p{2cm}|>{\centering}p{2cm}>{\centering}p{2cm}|cccc}
				\Xhline{4\arrayrulewidth}
				\multirow{2}{*}{\#} & \multicolumn{2}{c|}{\hspace{0.5cm}Uncertainty Reasoning} & \multirow{2}{*}{PCVM} & \multicolumn{2}{c|}{\hspace{0.2cm}Uncertainty-aware Loss} &  \cellcolor{red!7}  & \cellcolor{red!7}  &  \cellcolor{red!7}  & \cellcolor{red!7}  \\
				& Binary & Weighted && Masking & Reweighting& \multirow{-2}{*}{\cellcolor{red!7} Abs Rel} &\multirow{-2}{*}{\cellcolor{red!7} Sq Rel}  &  \multirow{-2}{*}{\cellcolor{red!7} RMSE}  & \multirow{-2}{*}{\cellcolor{red!7} RMSE log} \\
				\hline
				\hline
				1& consistency mask \cite{manydepth} & & & \ding{51}& &0.107&1.058 &5.934 &0.159 \\
				2& consistency mask \cite{manydepth} & & \ding{51}& \ding{51}& &0.103&0.953 &5.832 & 0.159 \\
				\hline
				3& segmentation mask \cite{dynamicdepth} && & \ding{51} &  &0.100 &0.961&5.620&0.150 \\
				4& segmentation mask \cite{dynamicdepth} && \ding{51}& \ding{51}    &&0.101 &0.965&5.647&0.150 \\
				\hline
				5& $[U>\gamma]$ &  &   &  \ding{51}  & &0.101&0.967&5.687&0.151 \\
				6&$[U>\gamma]$ & & \ding{51} & \ding{51}  &  &0.099&0.944 &5.616 & 0.151 \\
				
				7&  & $U$ & &\ding{51}&\ding{51} &0.100    &0.964 &5.630&0.151 \\
				8& & $U$  & \ding{51} & \ding{51} & &0.098  &0.903 &5.551&0.148 \\
				9& & $U$   & \ding{51} & & \ding{51} &0.097 &0.894&5.512 &0.146 \\
				10& & $U$   & \ding{51} & \ding{51} & \ding{51} &\textbf{0.095} &\textbf{0.882}&\textbf{5.490}&\textbf{0.146} \\
				\Xhline{4\arrayrulewidth}
			\end{tabular}%
		}
	\end{table}

	\noindent\textbf{PCVM.}\hspace{0.3cm}Our proposed PCVM module performs uncertainty-aware adaptive fusion of single-frame and multi-frame cues to modulate the misaligned matching cost distribution in the cost volume. The depth prediction performance is enhanced by PCVM in conjunction with the consistency mask~\cite{manydepth} and our uncertainty (row \#2,6). However, when incorporating a segmentation mask, performance degrades upon adding PCVM (row \#4). This decline is attributed to static objects included in segmentation masks, where only single-frame cues are utilized for those areas, while useful multi-frame cues are abandoned. Notably, with our weighted uncertainty representation, PCVM achieves a substantial performance improvement, reducing the absolute relative error from 0.100 to 0.095 (row \#7,10). This demonstrates that probabilistic fusion of single-frame and multi-frame cues is more effective than selecting one of them based on the binary criterion (row \#5,6). 
	Additionally, we present qualitative results with and without the PCVM module in Fig.~\ref{fig:5} (row \#7,10). The depth candidate $d_\text{low}$ with the lowest cost (\textit{i.e.}, highest probability) for each pixel in the cost volume, and the final prediction map are provided. In the case of dynamic objects, our PCVM effectively modulates the depth probability distribution in the cost volume by integrating single-frame cues, resulting in accurate depth predictions. In contrast, the model without PCVM propagates the erroneous distribution from the cost volume to the final prediction, leading to severe errors in dynamic areas.
	
	\noindent\textbf{Uncertainty-aware loss reweighting strategy.}\hspace{0.3cm}To address the incorrect self-supervision in dynamic areas, existing methods use a binary mask to exclude computed losses in those areas. In contrast, we propose a loss reweighting strategy, which reduces the computed loss based on the inferred probability (row \#9). Comparing our reweighting approach (row \#9) with binary masking (row \#8), we observe improved performance with the reweighting strategy. This is because binary masking lacks consideration of the detailed probability of areas being dynamic, and areas with ambiguous probability may not be adequately addressed due to thresholding. Furthermore, combining both masking and reweighting strategies yields the best performance (row \#10). This approach effectively excludes incorrect supervision in areas with high uncertainty through binary masking, while partially reducing the risk of providing incorrect supervision in remaining but potential dynamic areas according to the probability.

	\subsection{Generalization Study}
	We further validate the generalization ability of the proposed ProDepth and related works~\cite{dynamicdepth, manydepth} on the Waymo Open dataset~\cite{sun2020scalability}, which encompasses numerous dynamic objects and challenging scenes like low-light conditions during nighttime. We use 202 test video sequences for evaluation. The models are pretrained on the Cityscapes dataset. Given that DynamicDepth~\cite{dynamicdepth} necessitates a pretrained semantic segmentation network during inference, we pre-compute masks using EfficientPS~\cite{mohan2021efficientps}, which is utilized in experiments on Cityscapes and KITTI datasets. As shown in Table~\ref{tab:waymo}, ProDepth achieves superior performance compared to related works, showcasing its effective generalization ability.

	\begin{table*}[ht!]
		\centering
		\caption{Generalization study on the Waymo Open dataset.}
		\label{tab:waymo}
		\resizebox{\textwidth}{!}{%
			\begin{tabular}{c|cc|ccccccc}
				\Xhline{4\arrayrulewidth}
				Method &  Test frames & Semantics & \cellcolor{red!7} Abs Rel  & \cellcolor{red!7} Sq Rel  &  \cellcolor{red!7} RMSE  & \cellcolor{red!7} RMSE log & \cellcolor{green!7} $\delta < 1.25 $ & \cellcolor{green!7} $\delta < 1.25^2$  & \cellcolor{green!7} $\delta < 1.25^3$  \\
				\hline
				ManyDepth~\cite{manydepth}  & 2 (-1, 0) & & 0.260 & 3.916  &10.463 &0.313&0.606&0.856&0.941  \\
				DynamicDepth~\cite{dynamicdepth} & 2 (-1, 0) &\ding{51} & 0.255  & 3.521 & 9.902 & 0.313& 0.601& 0.856&0.942  \\
				\cellcolor{blue!7}
				\textbf{ProDepth} &2 (-1, 0) &  & \textbf{0.247}  &\textbf{3.462} &\textbf{9.544}&\textbf{0.300}&\textbf{0.628}&\textbf{0.873}&\textbf{0.949} \\
				\Xhline{4\arrayrulewidth}
			\end{tabular}%
		}
	\end{table*}
	\section{Conclusion}
	\label{sec:con}
	We present ProDepth, a multi-frame depth estimation framework addressing the inconsistency problem caused by dynamic objects in a probabilistic manner. Our contributions involve novel approaches of discerning the probability of areas being dynamic, direct rectification of misaligned cost volume with adaptive fusion of single-frame and multi-frame cues, and alleviating incorrect self-supervision in potential dynamic areas with a loss reweight strategy. ProDepth achieves state-of-the-art performance on both Cityscapes and KITTI datasets, and extensive experiments demonstrate the effectiveness of the proposed method.

	\noindent\textbf{Acknowledgement.}
	This work was supported by the Yonsei Signature Research Cluster Program of 2024 (2024-22-0161) and Institute of Information \& communications Technology Planning \& Evaluation (IITP) grant funded by the Korea government(MSIT) (No.2021-0-02068, Artificial Intelligence Innovation Hub).

	\clearpage
	\appendix
	\renewcommand\thesection{\Alph{section}}
	\renewcommand\thesubsection{\thesection.\arabic{subsection}}
	\noindent\Large{\textbf{Supplementary Materials}}
	\normalsize
	\section{Overview}
	\label{sec:a}
	This supplementary document provides additional technical details, experiments and visualization results. In Sec.~\ref{sec:b}, we describe implementation details of our ProDepth including hyperparameters and training strategies. In Sec.~\ref{sec:c}, we provide additional ablation study on the components of ProDepth and quantitative comparisons with related works. In Sec.~\ref{sec:d}, we discuss the limitations of our work. In Sec.~\ref{sec:e}, we present additional visualizations for diverse scenes.
	
	\section{Implementation Details}
	\label{sec:b}
	\noindent\textbf{Training.}\hspace{0.3cm}We implement our model in Pytorch~\cite{pytorch} with two NVIDIA RTX A6000 GPUs. Following the methodology in~\cite{manydepth}, we apply color and flip augmentations to training images. Unless explicitly specified, our models take two frames $\{I_{t-1},I_t\}$ as inputs during both training and testing, and three frames $\{I_{t-1},I_t,I_{t+1}\}$ are used for self-supervised training. The model undergoes training for 25 epochs on Cityscapes with batch size 24 and 20 epochs on KITTI with batch size 12. We employed the Adam optimizer~\cite{adam} with an initial learning rate of $10^{-4}$, reduced by a factor of 10 during the final 10 epochs for Cityscapes and 5 epochs for KITTI. Pose and single-frame networks are frozen when the learning rates drop. The loss coefficients are $\lambda_1=1$, $\lambda_2=0.3$, $\lambda_3=0.05$, and $\lambda_s=0.003$.
	
	\medskip
	\noindent\textbf{Model.}\hspace{0.3cm}The pose network uses ResNet18~\cite{resnet} as an encoder, while the depth network adopts a lightweight CNN-Transformer hybrid encoder from \cite{litemono}. In accordance with prior works, encoders are initialized with ImageNet~\cite{imagenet} pretrained weights. The features employed in constructing the cost volume have a channel size of $C=64$, with $k=128$ hypothesized depth bins (candidates), and a binary masking threshold of $\gamma = 0.8$.
	
	\medskip
	\noindent\textbf{Dataset.}\hspace{0.3cm}In our study of the Cityscapes dataset, we use a set of pre-processed 58,335 training images provided by \cite{dynamicdepth}, along with 1,525 images for testing. 
	For the KITTI dataset, we adhere to the Eigen split~\cite{eigen-split} following established practices~\cite{manydepth, depthformer, dynamicdepth, dualrefine}. This split encompasses 39,810 training images, 4,424 validation images, and 697 test images. For the generalization study on the Waymo Open dataset~\cite{sun2020scalability}, 2,216 front camera images are uniformly sampled from the validation set, which comprises 202 video sequences. In all datasets, we exclusively use unlabeled video frames, without incorporating additional segmentation masks or optical flow information. The ground-truth depth information is employed solely for evaluation, and we constrain the predicted depth values to be below 80 meters.
	
	\section{Additional Experimental Results}
	\label{sec:c}
	As outlined in the main paper, our experiments primarily concentrate on the Cityscapes dataset, which features a higher number of moving objects compared to the KITTI dataset. Unless otherwise specified, all experimental results denote performance on Cityscapes.

	\subsection{Fusion Method for Probabilistic Cost Volume Modulation}
\begin{figure*}[t]
		\centering  
		\includegraphics[width=0.9\linewidth]{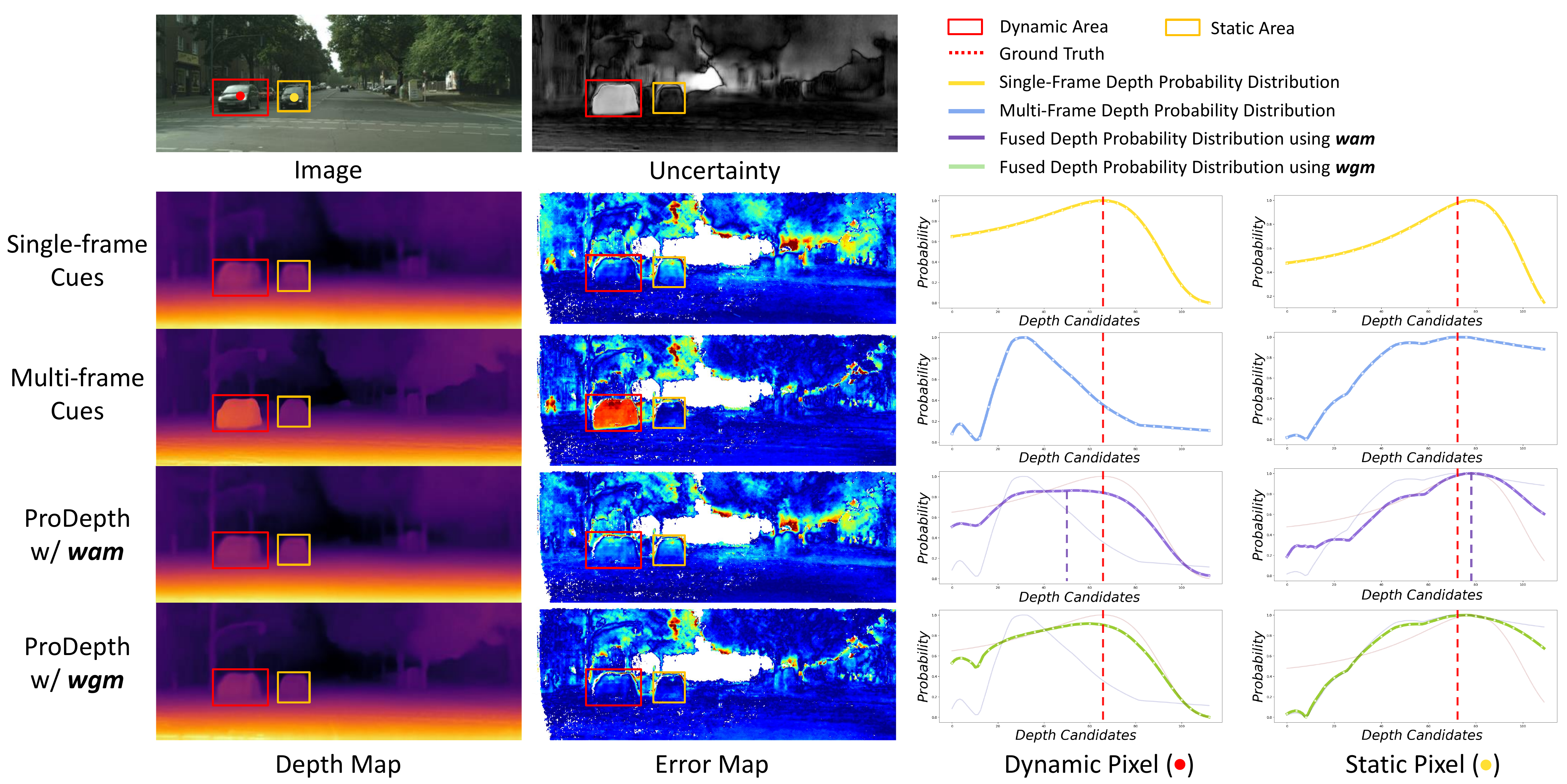}
		\caption{\textbf{Analysis on the fusion methods.} The estimated depth maps, error maps, and depth probability distributions are presented. Our proposed PCVM performs uncertainty-aware adaptive fusion of probability distributions derived from single- and multi-frame cues. When the weighted arithmetic mean (\textit{wam}) is used for fusion, the peak of the fused distribution exists between those in single- and multi-frame distributions and weighted geometric mean (\textit{wgm}). In contrast, when \textit{wgm} is used for fusion, the peak of the fused distribution follows that of more reliable cues according to the inferred uncertainty.}
		\label{fig:fusion}
	\end{figure*}

	In the proposed PCVM module, we perform an uncertainty-aware adaptive fusion of the depth probability distributions derived from single-frame and multi-frame cues in the cost volume. We explore weighted arithmetic mean (\textit{wam}) and weighted geometric mean (\textit{wgm}) as fusion methods. Given the probabilities $p_j \in \{p_{\text{single}}, p_{\text{cv}}\}$ and corresponding weights $w_j \in \{U, 1-U\}$, the fused probability distribution $P(d|x)$ can be obtained using \textit{wam} (Eq.~\ref{eq:wam}) or \textit{wgm} (Eq.~\ref{eq:wgm2}).
	
	\begin{align}
		\label{eq:wam}
		P(d|x) & = \frac{\sum_{j}{(p_j(d|x)\cdot{w_j}})}{\sum_{j}{w_j}}  =p_{\text{single}}(d|x)\cdot{U(x)} + p_{\text{cv}}(d|x)\cdot(1-U(x)). \\ 
		\label{eq:wgm2}
		P(d|x) & = (\prod_{j}{p_j(d|x)^{w_j}})^{1/{\sum_{j}{w_j}}} =  p_{\text{single}}(d|x)^{U(x)} \cdot p_{\text{cv}}(d|x)^{1-U(x)}.
	\end{align}
	As discussed in the main paper, the commonly used \textit{wam}, with its additive nature, may not guarantee the preservation of depth candidates at the maximum due to the linear combination of distributions. It tends to alter the location of a peak (local maxima) of the distribution after fusion, where the depth candidate with the highest probability in the fused probability distribution $P(d|x)$ does not precisely represent either single-frame or multi-frame cues. However, we observe that it is more appropriate to decisively adopt one position because in most cases, the multi-frame cue is more accurate than the single-frame cue in static scenes, and vice versa in dynamic scenes. Incorporating less reliable cue with \textit{wam} may shifts the positions of peaks away from the optimal depth candidate. In contrast, \textit{wgm} allows for the retention of depth candidates with the highest probability due to its multiplicative nature, maintaining the positions of peaks. Instead, their probabilities are adjusted with the corresponding weights. Table~\ref{tab:fusion} demonstrates that \textit{wam} degrades the performance, while \textit{wgm} achieves superior results. Fig.~\ref{fig:fusion} illustrates the analysis on the fusion methods.
	
	\begin{table*}[h]
		\caption{Fusion methods for PCVM.}
		\vspace{-9mm}
		\begin{center}
			\begin{adjustbox}{width=1\textwidth}
				\begin{tabular}{c|ccccccc}
					\Xhline{4\arrayrulewidth}
					Fusion Method  & \cellcolor{red!7} Abs Rel  & \cellcolor{red!7} Sq Rel  &  \cellcolor{red!7} RMSE  & \cellcolor{red!7} RMSE log & \cellcolor{green!7} $\delta < 1.25 $ & \cellcolor{green!7} $\delta < 1.25^2$  & \cellcolor{green!7} $\delta < 1.25^3$  \\
					\hline
					Weighted Arithmetic Mean& 0.098 & 0.945&5.715 &0.152&0.898&0.974&0.992  \\
					Weighted Geometric Mean & \textbf{0.095}  & \textbf{0.882}& \textbf{5.549}& \textbf{0.146} & \textbf{0.908}& \textbf{0.978}& \textbf{0.993}  \\
					\Xhline{4\arrayrulewidth}
				\end{tabular}
			\end{adjustbox}
			\label{tab:fusion}
		\end{center}
	\end{table*}

	\newpage
	\subsection{Depth Evaluation on Dynamic Objects}
	To validate the effectiveness of our approach, we further evaluate the model's performance on dynamic objects using the Cityscapes and Waymo Open datasets.
	
	\medskip
	\noindent\textbf{Cityscapes Dataset.}\hspace{0.3cm}
	For Cityscapes dataset, we compute the depth errors within
	movable objects belonging to dynamic classes (\eg, vehicles, pedestrians, bikes) as presented in Table~\ref{tab:dynamic}. These objects are identified using a pretrained semantic segmentation network.
	While DynamicDepth~\cite{dynamicdepth} and InstaDM~\cite{instadm} utilize these segmentation masks directly in both training and inference, our ProDepth achieves the comparable performance, underscoring the effectiveness of uncertainty reasoning and probabilistic cost volume modulation. It is important to note that the evaluation involves the static objects, as segmentation does not account for their movements.
	
	\begin{table*}[b]
		\caption{Depth errors on movable objects in dynamic classes.}
		\vspace{-9mm}
		\begin{center}
			\begin{adjustbox}{width=1\textwidth}
				\begin{tabular}{l|cc|ccccccc}
					\Xhline{4\arrayrulewidth}
					Method &   Semantics & $W \times H$ &\cellcolor{red!7} Abs Rel  & \cellcolor{red!7} Sq Rel  &  \cellcolor{red!7} RMSE  & \cellcolor{red!7} RMSE log & \cellcolor{green!7} $\delta < 1.25 $ & \cellcolor{green!7} $\delta < 1.25^2$  & \cellcolor{green!7} $\delta < 1.25^3$  \\
					\hline
					Monodepth2~\cite{monodepth2} & & $416 \times 128$ & 0.159& 1.937 &6.363 &0.201& 0.816 &0.950 &0.981 \\
					InstaDM~\cite{instadm}& \ding{51}&$832 \times 256$ &0.139& 1.698 &5.760& 0.181& 0.859 &0.959& 0.982 \\
					ManyDepth~\cite{manydepth}  &  & $416 \times 128$ & 0.169 & 2.175 &6.634& 0.218 &0.789 &0.921& 0.969  \\
					DynamicDepth~\cite{dynamicdepth} & \ding{51} & $416 \times 128$ & 0.129 &1.273 &4.626& \textbf{0.168}& \textbf{0.862}& \textbf{0.965} &0.986  \\
					\textbf{ProDepth w/o PCVM} & & $416 \times 128$ & 0.134& 1.151 &4.715 &0.177& 0.833 &0.958 &0.987   \\
					\cellcolor{blue!7}\textbf{ProDepth} & &$416 \times 128$  & \textbf{0.126}  &\textbf{0.953} &\textbf{4.483}&0.172&0.837&0.959&\textbf{0.988} \\
					\Xhline{4\arrayrulewidth}
				\end{tabular}
			\end{adjustbox}
			\label{tab:dynamic}
		\end{center}
	\end{table*}
	
	\begin{table*}[t]
		\caption{Generalization performance on static and dynamic areas in scenes involving moving objects.}
		\vspace{-8mm}
		\begin{center}
			\begin{adjustbox}{width=1\textwidth}
				\begin{tabular}{c|l|c|ccccccc}
					\Xhline{4\arrayrulewidth}
					Eval &  Method & Semantics &\cellcolor{red!7} Abs Rel  & \cellcolor{red!7} Sq Rel  &  \cellcolor{red!7} RMSE  & \cellcolor{red!7} RMSE log & \cellcolor{green!7} $\delta < 1.25 $ & \cellcolor{green!7} $\delta < 1.25^2$  & \cellcolor{green!7} $\delta < 1.25^3$  \\
					\hline
					\multirow{3}{*}{\rotatebox{90}{Static}} 
					&ManyDepth~\cite{manydepth}  &   & 0.259 & 3.770 &10.018& 0.320 &0.590 &0.849& 0.932  \\
					&DynamicDepth~\cite{dynamicdepth} & \ding{51} & 0.256 &3.634 &9.904& 0.321& 0.592& 0.849 &0.933  \\
					&\cellcolor{blue!7}\textbf{ProDepth} &   & \textbf{0.247}  &\textbf{3.626} &\textbf{9.483}&\textbf{0.299}&\textbf{0.634}&\textbf{0.863}&\textbf{0.936} \\
					\hline
					\multirow{3}{*}{\rotatebox{90}{Dynamic}} 
					&ManyDepth~\cite{manydepth}  &   & 0.376 & 6.661 &11.559& 0.381 &0.498 &0.757& 0.879  \\
					&DynamicDepth~\cite{dynamicdepth} & \ding{51} & 0.362 &6.100 &11.159& 0.363& 0.494& 0.773 &\textbf{0.900}  \\
					&\cellcolor{blue!7}\textbf{ProDepth} &   & \textbf{0.338}  &\textbf{5.976} &\textbf{11.088}&\textbf{0.346}&\textbf{0.553}&\textbf{0.797}&0.898 \\
					\Xhline{4\arrayrulewidth}
				\end{tabular}
			\end{adjustbox}
			\label{tab:waymo2}
			\vspace{-6mm}
		\end{center}
	\end{table*}
	
	\medskip
	\noindent\textbf{Waymo Open Dataset.}\hspace{0.3cm}
	As the Waymo Open dataset provides panoptic labels and 3D box positions, moving objects can be distinguished from static objects by computing their motions. We derive masks for moving objects following the procedure outlined in \cite{sun2024dynamo}, and then sample dynamic scenes containing at least one moving object. Table~\ref{tab:waymo2} presents the generalization performance on static and moving pixels within dynamic scenes. Our ProDepth model surpasses related approaches, benefiting significantly from PCVM, which compensates for the errors of multi-frame depth in dynamic areas. It is evident that PCVM significantly enhances performance in dynamic pixels compared to static pixels.

	\subsection{Additional Quantitative Results}
	\noindent\textbf{Predictive distribution for single-frame depth estimation.}\hspace{0.3cm}
	The predictive distribution can be modeled as Laplace or Gaussian. As shown in Table~\ref{tab:single}, the single-frame depth represented as a Gaussian distribution slightly outperforms the Laplace distribution in conveying useful cues for probabilistic fusion in a PCVM module.
	
	\begin{table*}[h]
		\caption{Predictive distribution for single-frame depth estimation.}
		\vspace{-7mm}
		\begin{center}
			\begin{adjustbox}{width=1\textwidth}
				\begin{tabular}{c|ccccccc}
					\Xhline{4\arrayrulewidth}
					Predictive Distribution  & \cellcolor{red!7} Abs Rel  & \cellcolor{red!7} Sq Rel  &  \cellcolor{red!7} RMSE  & \cellcolor{red!7} RMSE log & \cellcolor{green!7} $\delta < 1.25 $ & \cellcolor{green!7} $\delta < 1.25^2$  & \cellcolor{green!7} $\delta < 1.25^3$  \\
					\hline
					Laplace& 0.096 & 0.883  &5.579 &\textbf{0.146}&0.907&\textbf{0.978}&\textbf{0.993}  \\
					Gaussian & \textbf{0.095}  & \textbf{0.882}& \textbf{5.549}& \textbf{0.146} & \textbf{0.908}& \textbf{0.978}& \textbf{0.993}  \\
					\Xhline{4\arrayrulewidth}
				\end{tabular}
			\end{adjustbox}
			\label{tab:single}
		\end{center}
	\end{table*}
	
	\medskip
	\noindent\textbf{Binary masking threshold $\gamma$.}\hspace{0.3cm}
	Our uncertainty-aware photometric reprojection loss $\mathcal{L}_{up}$ consists of two factors: binary masking $M$ and loss reweighting $(1-U)$:
	\begin{equation}
		\label{eq:up}
		\mathcal{L}_{up} = M\odot(1-U)\odot\mathcal{L}_{p},\hspace{0.3cm}M = [U < \gamma],
	\end{equation}
	where $\odot$ is element-wise product and $[\cdot]$ denotes the Iverson bracket. In Table~\ref{tab:gamma}, we present the results obtained with various thresholds for binary masking. We adopt $\gamma = 0.8$ for the final model, which excludes dynamic areas with high uncertainty ($U > 0.8$).
	
	\begin{table*}[h]
		\caption{Ablation on the binary masking threshold $\gamma$.}
		\vspace{-3mm}
		\begin{center}
			\begin{adjustbox}{width=0.8\textwidth}
				\begin{tabular}{c|ccccccc}
					\Xhline{4\arrayrulewidth}
					Threshold $\gamma$  & \cellcolor{red!7} Abs Rel  & \cellcolor{red!7} Sq Rel  &  \cellcolor{red!7} RMSE  & \cellcolor{red!7} RMSE log & \cellcolor{green!7} $\delta < 1.25 $ & \cellcolor{green!7} $\delta < 1.25^2$  & \cellcolor{green!7} $\delta < 1.25^3$  \\
					\hline
					0.2& 0.101 & 0.978 &5.781 &0.153&0.898&0.975&0.992  \\
					0.4& 0.096 & 0.883  &5.595 &0.148&0.904&0.977&0.992  \\
					0.6 & \textbf{0.095} & \textbf{0.869}  &5.598& 0.148 &0.904&0.977&\textbf{0.993}  \\
					0.8& \textbf{0.095}  & 0.882& \textbf{5.549}& \textbf{0.146} & \textbf{0.908}& \textbf{0.978}& \textbf{0.993}  \\
					\Xhline{4\arrayrulewidth}
				\end{tabular}
			\end{adjustbox}
			\label{tab:gamma}
		\end{center}
		\vspace{-3mm}
	\end{table*}
	
	\medskip
	\noindent\textbf{KITTI evalution on improved ground truth.}\hspace{0.3cm}
	In Table~\ref{tab:kitti}, we present the KITTI results evaluated using the improved dense ground truth~\cite{uhrig2017sparsity}, which is generated by accumulating 5 consecutive frames to form a denser ground truth depth map. Our approach exhibits comparable performance to the supervised method BTS~\cite{lee2019big}, showcasing the effectiveness of our self-supervised multi-frame framework.
	
	\begin{table*}[h]
		\caption{Depth evaluation on the KITTI dataset using the improved ground truth depth maps. \textit{D} indicates the depth supervision and \textit{M} denotes the monocular self-supervision.}
		\vspace{-4mm}
		\begin{center}
			\begin{adjustbox}{width=1\textwidth}
				\begin{tabular}{l|cc|ccccccc}
					\Xhline{4\arrayrulewidth}
					\multirow{2}{*}{Method}&\multirow{2}{*}{Supervision} & \multirow{2}{*}{Test frames}  & \multicolumn{4}{c}{\cellcolor{red!7}Error metric ($\downarrow$)}    & \multicolumn{3}{c}{ \cellcolor{green!7}Accuracy metric ($\uparrow$)}  \\
					& & &   
					\cellcolor{red!7} Abs Rel & \cellcolor{red!7} Sq Rel&  \cellcolor{red!7} RMSE  & \cellcolor{red!7} RMSE log & \cellcolor{green!7} $\delta < 1.25 $ & \cellcolor{green!7} $\delta < 1.25^2$  & \cellcolor{green!7} $\delta < 1.25^3$     \\
					\hline				
					Kuznietsov \etal~\cite{kuznietsov2017semi} & D & 1 & 0.113& 0.741& 4.621& 0.189& 0.862 &0.960& 0.986 \\
					Gan \etal~\cite{gan2018monocular}& D & 1 & 0.098& 0.666 &3.933 &0.173& 0.890& 0.964& 0.985\\
					Guizilimi \etal~\cite{guizilini2020robust} & D & 1 & 0.072& 0.340& 3.265& 0.116& 0.934 & - & -\\
					DORN~\cite{fu2018deep} & D & 1 & 0.072& 0.307& 2.727& 0.120& 0.932& 0.984& 0.994\\
					Yin \etal~\cite{yin2019enforcing} & D & 1 & 0.072 &- & 3.258& 0.117& 0.938 &0.990& 0.998\\
					BTS~\cite{lee2019big} & D & 1 & 0.059& 0.245& 2.756& 0.096 &0.956& 0.993& 0.998\\
					
					\hline
					Johnston \etal~\cite{johnston2020self}&M & 1 &   0.081 & 0.484  & 3.716  & 0.126  & 0.927  & 0.985  & 0.996 \\

					Packnet-SFM~\cite{packnet}&M & 1 &0.078 & 0.420 &  3.485 &  0.121 &  0.931 &  0.986 &  0.996\\
					
					Monodepth2~\cite{monodepth2}&M & 1 & 0.090 & 0.545 & 3.942 & 0.137 & 0.914 & 0.983 & 0.995\\

					\arrayrulecolor{gray}\cline{2-10}
					Patil \etal~\cite{patil2020dont}&M & N  & 0.087  &   0.495  &   3.775  &   0.133  &   0.917  &   0.983  &   0.995  \\
					Wang \etal~\cite{wang2020self} &M& 2 (-1, 0) &  0.082 & 0.462 & 3.739 & 0.127 &  0.923 &  0.984 & 0.996\\
					ManyDepth~\cite{manydepth}&M & 2 (-1, 0) &  0.070  &0.399  &3.455  &0.113  &0.941  &0.989  &0.997\\
					DynamicDepth~\cite{dynamicdepth}&M & 2 (-1, 0) &  0.068  &   0.362  &  3.454  &  0.111  &  0.943 &0.991  &  \textbf{0.998}\\
					\cellcolor{blue!7}
					\textbf{ProDepth}& M& 2 (-1, 0)   & \textbf{0.059} & \textbf{0.308 }& \textbf{3.060}& \textbf{0.097}& \textbf{0.959}& \textbf{0.992}& 0.997 \\
					\Xhline{4\arrayrulewidth}
				\end{tabular}
			\end{adjustbox}
			\label{tab:kitti}
		\end{center}
	\end{table*}
	
	\newpage
	
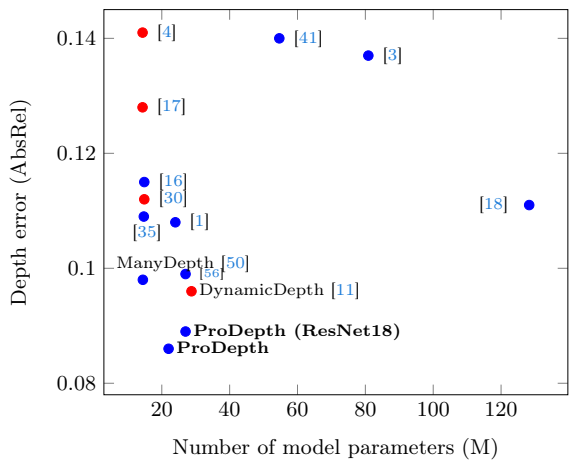
\begin{figure*}[t]
		\centering
		\begin{tikzpicture}[scale=0.9]
			\begin{axis}[
				xlabel = Number of model parameters (M),
				ylabel = Depth error (AbsRel),
				ymin = 0.078,
				ymax = 0.145,
				scaled y ticks=false,
				tick label style={/pgf/number format/fixed}]
				\addplot[
				scatter/classes={b={blue}, r={red}},
				scatter, mark=*, only marks, 
				scatter src=explicit symbolic,
				nodes near coords*={\Label},
				visualization depends on={value \thisrow{label} \as \Label} ,
				visualization depends on={value \thisrow{anchor}\as\myanchor},
				every node near coord/.append style={anchor=\myanchor},
				] table [meta=class] {
					x y class label anchor
					14.34 0.141 r {\scriptsize \cite{casser2019depth}} west
					80.88 0.137 b {\scriptsize\cite{bian2019unsupervised}} west
					54.64 0.140 b \scriptsize{\cite{ranjan2018adversarial}} west
					14.33 0.128 r {\scriptsize\cite{gordon2019depth}} west
					14.84 0.115 b {\scriptsize\cite{monodepth2}} west
					14.84 0.112 r {\scriptsize\cite{instadm}} west
					14.7 0.109 b {\scriptsize\cite{hrdepth}} north
					128.29 0.111 b {\hspace{-1cm}\scriptsize\cite{packnet}} east
					24 0.108 b {\scriptsize\cite{bae2023deep}} west
					27 0.099 b {\tiny\cite{zhao2022monovit}} {west}
					14.421 0.098 b {\hspace{1.2cm}\scriptsize ManyDepth~\cite{manydepth}} south
					28.75 0.096 r {\scriptsize DynamicDepth~\cite{dynamicdepth}} west
					22 0.086 b \scriptsize\textbf{ProDepth} west
					27 0.089 b \scriptsize\textbf{ProDepth (ResNet18)} west
				};
			\end{axis}
		\end{tikzpicture}
		\caption{Depth error on KITTI dataset against the number of model parameters. Red dots indicate models requiring semantics, and the parameters of segmentation network are not considered.}\label{fig:size}
	\end{figure*}

	\medskip
	\noindent\textbf{Model size and runtime.}\hspace{0.3cm}
	Figure~\ref{fig:size} illustrates the depth error on the KITTI dataset plotted against the number of model parameters. Our ProDepth achieves the best performance while maintaining a comparable number of parameters. When we adopt ResNet18~\cite{resnet} as the depth encoder, the performance slightly decreases while involving more parameters. ProDepth runs at 23FPS on a Titan RTX GPU.
	
	\section{Limitation}
	\label{sec:d}
	Our approach is grounded in the widely accepted observation~\cite{manydepth, dynamicdepth, depthformer, li2023learning, monorec} that single-frame-based prediction outperforms multi-frame-based prediction in dynamic areas. However, it is important to note that single-frame estimation might struggle to achieve accurate depth for moving objects, particularly for textureless or low-light pixels, and may not offer useful cues. In addition, enabling unsupervised single-frame depth learning for dynamic regions relies on transferring knowledge from static objects, which requires a careful training strategy. The training challenges posed by datasets containing an abundance of moving objects further complicate this process.
	
	\section{Additional Visualizations}
	\label{sec:e}
	
\begin{figure*}[t]
		\centering  
		\includegraphics[width=0.9\linewidth]{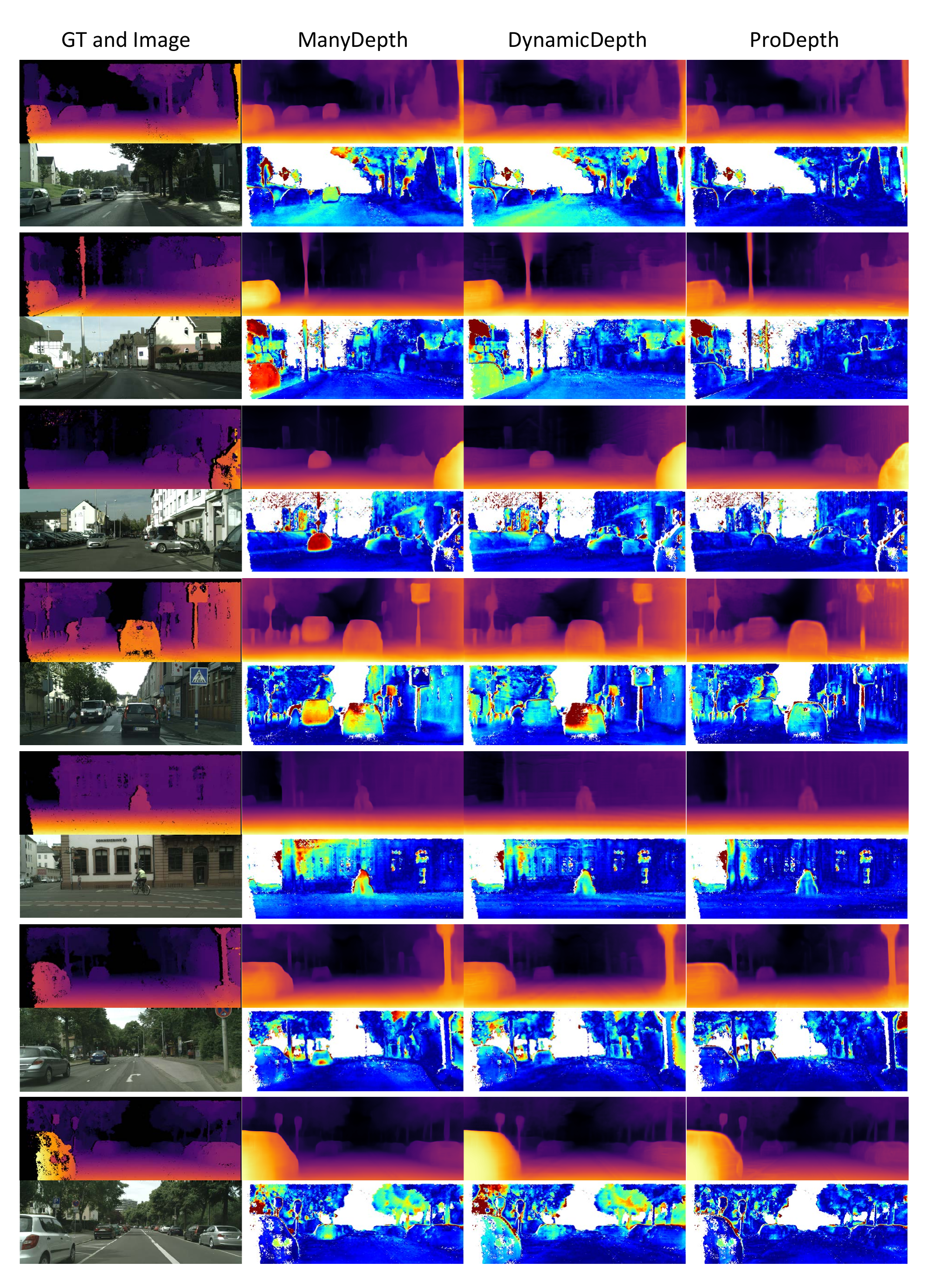}
		\caption{\textbf{Further qualitative results on the Cityscapes dataset (Part 1)}. Error maps in the second row for each scene measure the absolute relative error compared to the ground truth after median scaling \cite{eigen2015predicting}, depicting large errors in red and small errors in blue.}
		\label{fig:vis1}
	\end{figure*}

\begin{figure*}[t]
		\centering  
		\includegraphics[width=0.9\linewidth]{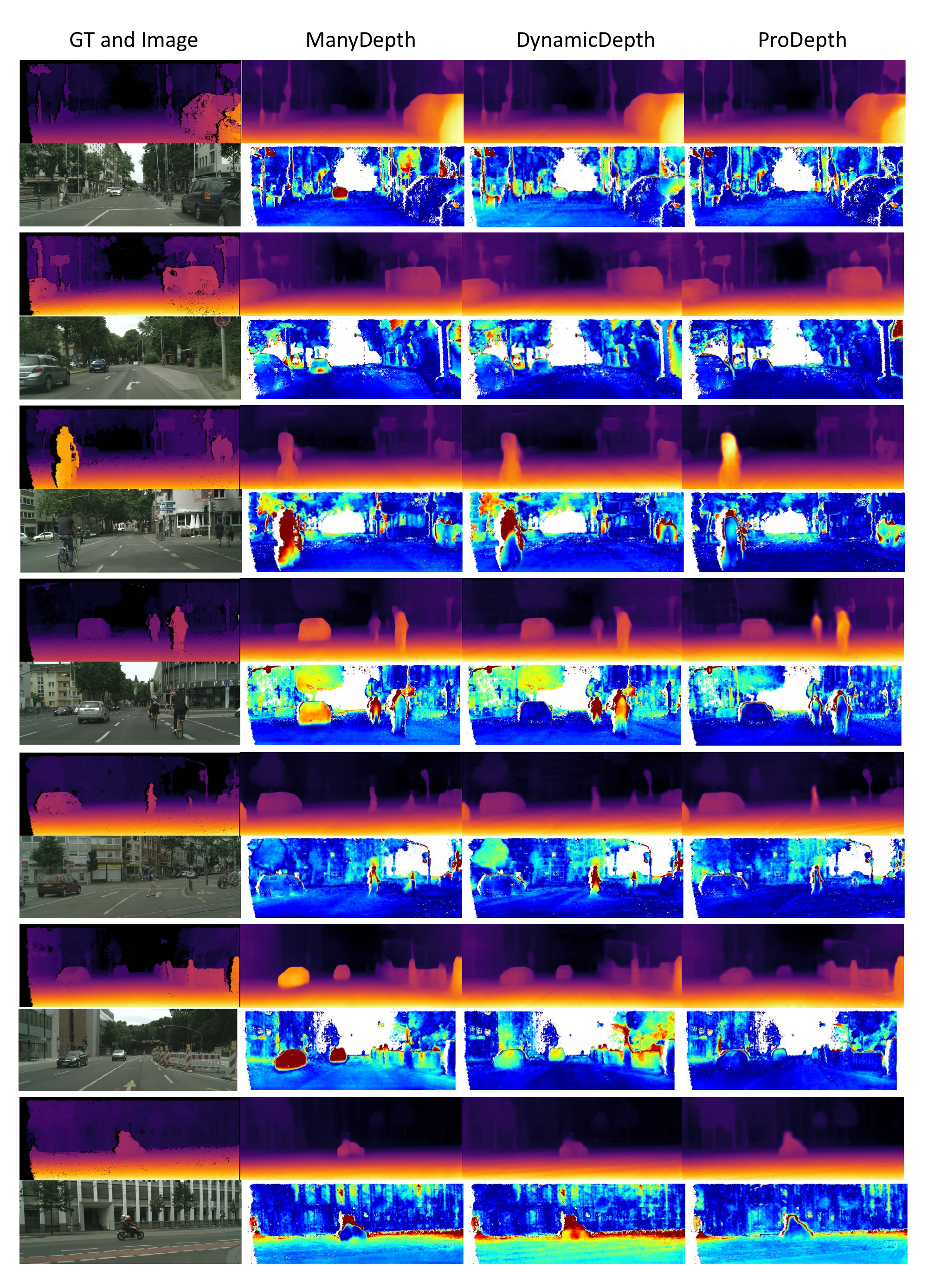}
		\caption{\textbf{Further qualitative results on the Cityscapes dataset (Part 2)}. Error maps in the second row for each scene measure the absolute relative error compared to the ground truth after median scaling \cite{eigen2015predicting}, depicting large errors in red and small errors in blue.}
		\label{fig:vis2}
	\end{figure*}

	We provide additional qualitative comparisons with related works~\cite{manydepth, dynamicdepth} in Figure~\ref{fig:vis1} and Figure~\ref{fig:vis2}. Our ProDepth demonstrates accurate depth estimation, particularly in dynamic areas, highlighting the effectiveness of our probabilistic approach.

	\clearpage
	
	\bibliographystyle{splncs04}
	\bibliography{main}
\end{document}